\documentclass[letterpaper]{article} 
 \usepackage{aaai25}
\usepackage{times}  
\usepackage{helvet}  
\usepackage{courier}  
\usepackage[hyphens]{url}  
\usepackage{graphicx} 
\usepackage{natbib}  
\usepackage{caption} 
\usepackage{algorithm}
\usepackage{algorithmic}
\usepackage{pifont}

\usepackage{amssymb}
\usepackage{amsmath}
\usepackage{multirow}
\usepackage{booktabs}
\usepackage{xcolor}

\usepackage{newfloat}
\usepackage{listings}
\DeclareCaptionStyle{ruled}{labelfont=normalfont,labelsep=colon,strut=off} 
\floatstyle{ruled}
\newfloat{listing}{tb}{lst}{}
\floatname{listing}{Listing}
\nocopyright

\title{O-Mamba: O-shape State-Space Model for Underwater Image Enhancement}
\author {
    Chenyu Dong\(^1\)\thanks{These authors contributed equally to this work.},
    Chen Zhao\(^2\)\(^*\),
    Weiling Cai\(^1\)\thanks{Corresponding Author},
    Bo Yang\(^1\),
}
\affiliations {
    \(^1\)School of Computer and Electronic Information, Nanjing Normal University, China\\
    \(^2\)School of Intelligent Science and Technology, Nanjing University, China\\
    dongcy@njnu.edu.cn, 2518628273@qq.com, caiwl@njnu.edu.cn, yangbo@nnu.edu.cn
}

\begin{document}

\maketitle

\begin{abstract}
Underwater image enhancement (UIE) face significant challenges due to complex underwater lighting conditions. Recently, mamba-based methods have achieved promising results in image enhancement tasks. However, these methods commonly rely on Vmamba, which focuses only on spatial information modeling and struggles to deal with the cross-color channel dependency problem in underwater images caused by the differential attenuation of light wavelengths, limiting the effective use of deep networks.
In this paper, we propose a novel UIE framework called O-mamba. O-mamba employs an O-shaped dual-branch network to separately model spatial and cross-channel information, utilizing the efficient global receptive field of state-space models optimized for underwater images.
To enhance information interaction between the two branches and effectively utilize multi-scale information, we design a Multi-scale Bi-mutual Promotion Module. This branch includes MS-MoE for fusing multi-scale information within branches, Mutual Promotion module for interaction between spatial and channel information across branches, and Cyclic Multi-scale optimization strategy to maximize the use of multi-scale information.
Extensive experiments demonstrate that our method achieves state-of-the-art (SOTA) results.The code is available at https://github.com/chenydong/O-Mamba.
\end{abstract}




\section{Introduction}

Underwater image enhancement(UIE) has become a critical area of research due to its wide range of applications, from underwater explorations \cite{9197308, liu2022novel} to underwater robotics \cite{9359465,liu2020underwater}. The unique challenges posed by underwater environments, such as varying light conditions, suspended particles, and the absorption and scattering of light, result in images that often suffer from low illumination \cite{hou2023non}, color distortion \cite{gao2019underwater}, and loss of detail \cite{wang2021leveraging}. These issues complicate the processing and analysis of underwater images, making enhancement techniques essential for ensuring the effectiveness of subsequent tasks.

To address the challenges posed by complex underwater environments, traditional UIE methods based on the physical characteristics of underwater images have been proposed \cite{drews2013transmission,peng2018generalization, peng2017underwater,akkaynak2018revised}. These methods heavily rely on various prior assumptions, physical models, and classical, non-data-driven image processing techniques \cite{cong2024comprehensivesurveyunderwaterimage,10.1007/978-981-99-7549-5_1}. However, due to the limited representational capacity of these physics-based methods and the inherent complexity of underwater scenes, it is difficult to achieve good enhancement results using a physical model for complex underwater conditions \cite{anwar2020diving}.

Recognizing the powerful representational capabilities of neural networks, several methods based on CNN were introduced to address the limitations of traditional methods in handling complex underwater scenes \cite{li2020underwater,mei2022uir,Naik_Swarnakar_Mittal_2021}. Although CNN-based methods have achieved notable results, they are limited by their restricted receptive fields and are therefore not effective at capturing global dependencies within underwater scenes\cite{He_2016_CVPR}. To better capture global information, Transformer-based methods have been introduced to the UIE field \cite{peng2023u,wen2024waterformer,huang2022underwater}. The self-attention mechanism excels in modeling long-range dependencies and effectively extract global features, but their quadratic computational complexity with respect to sequence length presents challenges when processing high-resolution  images. 

\begin{figure}[t]
\centering
\includegraphics[width=0.47 \textwidth]{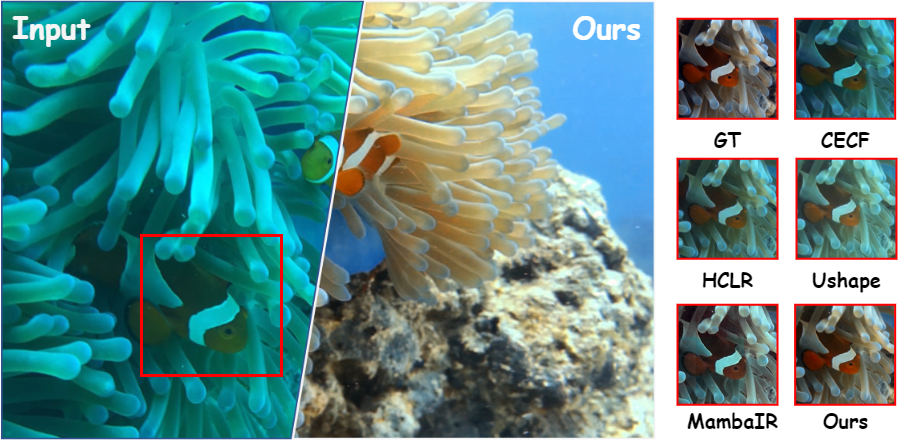}
\caption{
Visualization of high-resolution image. Compared to the latest SOTA methods, our method achieves the best color correction. Please zoom to view.}
\label{intro}
\end{figure}

Recently, 
state space model, such as Mamba \cite{gu2024mambalineartimesequencemodeling}, have garnered increasing attention due to their linear complexity and suitability for long sequences. These models offer a compelling alternative by effectively balancing the need for capturing long-range dependencies and global features while maintaining computational efficiency \cite{liang2021swinir}.
Existing Mamba-based methods \cite{liu2024vmambavisualstatespace,zhu2024visionmambaefficientvisual} have achieved significant results in various  image restoration fields. 
These 
methods \cite{guo2024mambairsimplebaselineimage,cheng2024activatingwiderareasimage,zheng2024ushapedvisionmambasingle} typically rely on standard Mamba blocks, which use 2D selective scanning modules with four spatial directions, limiting the capability of modeling channels representation. 
However, in UIE tasks, unique color degradation caused by the differential attenuation of light wavelengths \cite{berman2020underwater} is closely related to cross-channel information, fully leveraging channel information becomes crucial. These existing models lack sufficient attention to channel information, Resulting in insufficient color correction capability. As shown in Figure 1, the recent SOTA methods fail to remove the underwater color disturbances, appearing weaker color correction capability. 

In this paper, we propose a novel framework called O-Mamba for underwater image enhancement, which adopts an O-shaped dual-branch network that fully leverages the efficient global receptive fields provided by Spatial Mamba (SM) and Channel Mamba (CM) blocks to model spatial and cross-channel features separately.
To enable information interaction between the two branches and effectively utilize multi-scale information, we design a Multi-scale Bi-mutual Promotion (MSBMP) Module. This module consists of three components: 
1) To 
fully utilize multi-scale information within each branch, we introduce Multi-scale Mixture of Experts (MS-MoE), which uses multiple Mamba experts to learn feature representations at different scales.
2) 
The mutual Promotion (MP) module is designed, aiming to achieve fusion of spatial and channel information. 
3)Moreover,  we introduce the Cyclic Multi-scale (CMS) optimization strategy to  alleviate the interference from optimizing multiple scales simultaneously.
By leveraging the O-shaped dual-branch network and the MSBMP Module, O-Mamba achieves comprehensive attention to spatial and cross-channel information, effectively restoring underwater images by fully utilizing multi-scale information.

In summary, the main contributions of our method are as follows:

\begin{itemize}
    \item We propose a novel framework, namely O-Mamba, which is an O-shaped dual-branch network consisting of the spatial Mamba branch and the channel Mamba branch. 
    \item We design a Multi-scale Bi-mutual Promotion Module, to effectively manage the information interaction between the two branches of O-Mamba. This Module includes MS-MoE, MP module, and CMS optimization strategy.
    \item Extensive experiments have demonstrated that O-Mamba outperforms other UIE methods, achieving state-of-the-art (SOTA) performance.
\end{itemize}

\section{Related Works}

\subsection{Underwater Image Enhancement}

Current underwater image enhancement (UIE) methods can be broadly categorized into non-deep learning-based and deep learning-based approaches. Non-deep learning methods typically rely on prior assumptions, physical models, and classical image processing techniques. For example, \citeauthor{drews2013transmission} used the Dark Channel Prior (DCP) to estimate transmission maps in underwater environments, while \citeauthor{akkaynak2018revised} improved the atmospheric scattering model for underwater image restoration. \citeauthor{ABDULGHANI2017181} applied histogram-based methods to enhance visual appearance. However, these methods often suffer from inaccurate physical parameter estimation and are limited by their assumptions, making them less effective in diverse underwater scenes.

In contrast, as deep learning has been widely used in various fields and achieved great performance \cite{zhou2023pyramiddiffusionmodelslowlight,Lu_2024_CVPR,Zhou_2024_CVPR,Lu_2023_ICCV,zhou2024migcadvancedmultiinstancegeneration}, deep learning-based underwater image enhancement (UIE) methods have also emerged \cite{drews2013transmission, ABDULGHANI2017181, akkaynak2018revised, mei2022uir,10448182, peng2023u, cong2024underwater,zhao2024wavelet} . The introduction of new underwater image datasets like UIEB \cite{8917818} and LSUI \cite{peng2023u} has significantly accelerated the development of deep learning models. CNN-based models such as WaterNet \cite{8917818} and Ucolor \cite{Ucolor} have achieved end-to-end underwater image restoration, while Transformer-based architectures like Ushape \cite{peng2023u} and WaterFormer \cite{wen2024waterformer} have further improved restoration outcomes. However, CNNs struggle with modeling global dependencies due to their limited receptive fields, and although attention mechanisms  provide global receptive fields, they come with quadratic computational complexity, making efficient attention design challenging and often requiring a trade-off between computational efficiency and global modeling capabilities.

\begin{figure*}[t]
\centering
\includegraphics[width=1\textwidth]{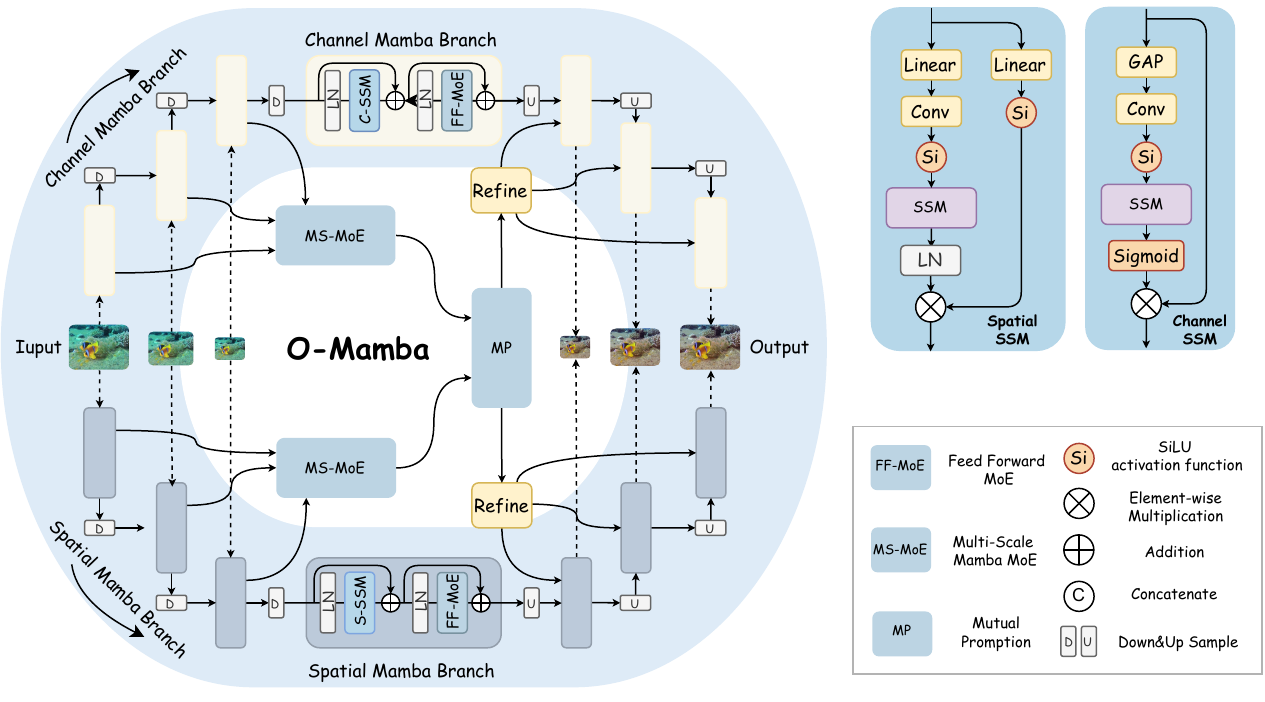}
\caption{Overall 
O-Mamba is an O-shaped dual-branch network consisting of a Spatial Mamba Branch and a Channel Mamba Branch. The Spatial Mamba Branch is designed to capture spatial dependencies in images, while the Channel Mamba Branch, utilizing the Channel Mamba Block, focuses on capturing cross-channel dependencies. The interaction between the two branches is achieved through the Multi-scale Bi-mutual Promotion Module, which consists of three parts: MS-MoE for integrating multi-scale information within branches, the Mutual Promotion module for integrating spatial and channel information between branches and the Cyclic Multi-scale (CMS) optimization strategy for optimize multi-scale losses.}
\label{fm}
\end{figure*}

\subsection{State Space Models}
State Space Models (SSMs) \cite{gu2022efficientlymodelinglongsequences,gu2021combining,nguyen2022s4nd}, derived from classical control theory \cite{kalman1960new}, have gained attention for their linear complexity and ability to model long sequences. The Structured State-Space Sequence model (S4) \cite{gu2022efficientlymodelinglongsequences} was one of the first to demonstrate the potential of SSMs, followed by the S5 layer \cite{smith2023simplifiedstatespacelayers}, which introduced MIMO SSM and efficient parallel scanning. Further advancements, such as the H3 model \cite{fu2023hungryhungryhipposlanguage} and the Gated State Space layer \cite{mehta2022longrangelanguagemodeling}, have narrowed the performance gap between SSMs and Transformers in natural language tasks. Mamba \cite{gu2023mamba}, a data-dependent SSM with a selective mechanism, has shown superior performance in both natural language processing and various vision tasks, including image classification and image restoration.
Mamba-based image restoration methods \cite{guo2024mambairsimplebaselineimage,cheng2024activatingwiderareasimage,zheng2024ushapedvisionmambasingle} typically rely on standard Mamba blocks, which use 2D selective scanning modules. These modules scan images from only four spatial directions, lacking the ability to scan channels which is important in UIE tasks. These existing models lack sufficient attention to channel information, resulting in inadequate channel modeling capabilities, making them ineffective for processing underwater images.

\section{Preliminaries}
State Space Models (SSMs), inspired by continuous linear time-invariant (LTI) systems, offer a robust framework for sequence-to-sequence modeling. These models use an implicit latent state \( h(t) \in \mathbb{R}^N \) to map a 1-dimensional function or sequence input \( x(t) \in \mathbb{R} \) to an output \( y(t) \in \mathbb{R} \). Mathematically, the system can be represented by the following linear ordinary differential equations (ODE):
\begin{equation}
\begin{aligned}
    h'(t) &= Ah(t) + Bx(t), \\
     y(t) &= Ch(t) + Dx(t)
\end{aligned}
\label{ssm1}
\end{equation}
In this context, \( N \) is the hidden state size, \( A \in \mathbb{R}^{N \times N} \), \( B \in \mathbb{R}^{N \times 1} \), \( C \in \mathbb{R}^{1 \times N} \) are the parameters for a state size, and \( D \in \mathbb{R} \) represents the skip connection.

To integrate Equation \ref{ssm1} into deep learning algorithms, the Zero-Order Hold (ZOH) rule is typically used for discretization. This rule, which involves the timescale parameter \(\Delta\), converts the continuous parameters \(A\) and \(B\) into their discrete counterparts \(\bar{A}\) and \(\bar{B}\). The definitions are as follows:

\begin{equation}
\begin{aligned}
    h'_t &= \bar{A} h_{t-1} + \bar{B} x_t, \\
    y_t &= C h_t + D x_t
\end{aligned}
\end{equation}

where the discretized parameters are defined as:

\begin{equation}
\begin{aligned}
    \bar{A} &= e^{\Delta A}, \\
    \bar{B} &= (\Delta A)^{-1}(e^{\Delta A} - I) \cdot \Delta B
\end{aligned}
\end{equation}

Recently, the introduction of Mamba \cite{gu2024mambalineartimesequencemodeling} has further advanced the capabilities of SSMs by incorporating a selective mechanism.
This enables Mamba to effectively process very long sequences while also utilizing parallel scanning algorithms for faster parallel computing, thereby achieving efficient training.

\section{Methodology}
\subsection{Overall Framework}

As shown in Figure \ref{fm}, our O-Mamba is an O-shaped network that consists of two primary branches: the Spatial Mamba Branch and the Channel Mamba Branch. The Spatial Mamba Branch is a UNet-like network composed of stacked Spatial Mamba Blocks. It enhances the spatial structure of the image by extracting spatial information at different scales, thereby improving the image's overall spatial integrity. In contrast, the Channel Mamba Branch is composed of stacked Channel Mamba Blocks, focusing on cross-channel degradation information at different scales.
To enable information interaction between the two branches and fully utilize multi-scale information, we design a Multi-scale Bi-mutual Promotion(MSBMP) Module. The middle section of Figure \ref{fm} represents MSBMP Module. It consists of the Multi-scale Mixture of Experts (MS-MoE) for intra-branch feature fusion, the Mutual Promotion (MP) module for inter-branch interaction, and a Cyclic Multi-scale (CMS) optimization strategy.

\subsection{Spatial Mamba Branch}
\subsubsection{Spatial Mamba Block.}
The spatial Mamba block is built following the fundamental architecture of Transformers. As shown in Fig \ref{fm}, We utilize a four-directional selective scan, namely the Spatial Selective Scan Module (S-SSM), to replace self-attention and capture long-term dependency comprehensively across the spatial dimension. Simply replacing the attention mechanism with SSM does not yield optimal results \cite{guo2024mambairsimplebaselineimage}. Inspired by \citeauthor{he2024frequency,pióro2024moemambaefficientselectivestate}, we design a Feed Forward Mixture of Experts (FF-MoE) to replace the MLP in Transformers, with further details to be explained later.
The input of SM block consists of two parts: one is the input of the previous block \(Y^{(n-1)}\),   another is the multi-scale input of the current scale \(X_{n}\), the process of the current block can be defined as follows:

\begin{equation}
\begin{aligned}
    X_{in} = X_{n}&+Y^{(n-1)}_s, \\
    X_{mid} = S\text{-}SSM(L&N(X_{in})) + X_{in}, \\
    Y^{(n)}_s = FF\text{-}MoE(L&N(X_{mid})) + X_{mid}
\end{aligned}
\end{equation}

Here, \(LN\) represents LayerNorm and n \( \in \{1,2,3\}\) represents the number of current mamba block. 

\subsubsection{Spatial Selective Scan Module.}
In the aspect of spatial information modeling, attention mechanisms often need to balance between computational efficiency and effective global receptive fields \cite{liang2021swinir}. Inspired by \citeauthor{liu2024vmambavisualstatespace}, we employ Spatial-SSM, which provides global modeling with linear complexity, to handle the global spatial information in underwater images. As illustrated, the input \( X \) of the Spatial-SSM is processed through two branches.
In the first branch, the feature channels are expanded to \( \lambda C \) via a linear layer, where \( \lambda \) is a predefined channel expansion factor. This is followed by a convolutional layer, a \(SiLU\) \cite{ELFWING20183} activation function, a \(SSM\) layer, and \(LayerNorm\). Specifically, this branch can be expressed as follows:

\begin{equation}
\begin{aligned}
    X_a = LN(SSM(SiLU(Conv(Linear(X))))),
\end{aligned}
\end{equation}

Similar to the first branch, second branch expand the feature channels to \( \lambda C \) followed by the SiLU activation function: 

\begin{equation}
\begin{aligned}
    X_b = SiLU(Linear(X))),
\end{aligned}
\end{equation}

Subsequently, the features from both branches are aggregated using the Hadamard \cite{horn1990hadamard} product.  The channel dimension is projected back to \( C \) to produce an output \( X_{out} \) that matches the shape of the input:

\begin{equation}
\begin{aligned}
    X_{sout} = Linear(X_a \odot X_b),
\end{aligned}
\end{equation}

where \( \odot \) represents the Hadamard product.

\begin{figure}[t]
\centering
\includegraphics[width=0.47\textwidth]{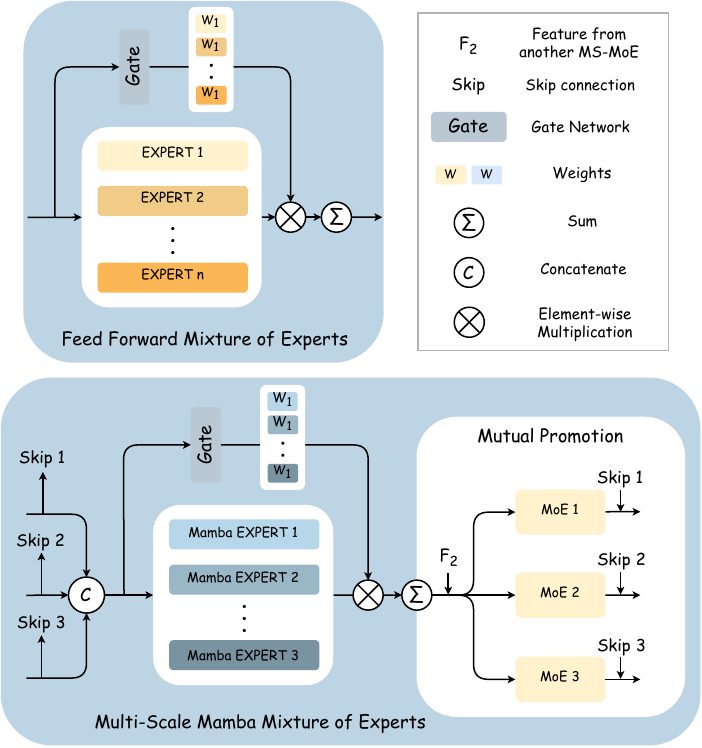}
\caption{The architecture of the Feed Forward MoE and Multi-scale Mamba MoE.}
\label{moe}
\end{figure}

\subsubsection{Feed Forward Mixture of Experts}
Given the diversity and complexity of underwater environments, including but not limited to blue, green, low-light, and turbid scenes, it is challenging for a single network to effectively model such a wide range of underwater images. Inspired by \citeauthor{he2024frequency,riquelme2021scaling}, we incorporate a Mixture of Experts (MoE) as the feed forward network, as shown in Figure \ref{moe}, enabling Mamba to handle various scenarios without increasing computational costs. In this setup, FFMoE acts as the experts at the current scale, dynamically selecting the most suitable expert through \(Gate\) network to address the current underwater conditions.
To define the FF-MoE process, we begin with the input \(X\). The process is as follows:

\begin{equation}
\begin{aligned}
     X_{out} = \sum_{i=1}^{N} &Expert_i(X_{}) \cdot Gate(X),
\end{aligned}
\end{equation}

\subsection{Channel Mamba Branch}
\subsubsection{Channel Mamba Block.}
Similar to the Spatial Mamba Block, the Channel Mamba Block employs another SSM to replace self-attention, which we refer to as the Channel Selective Scan Module (C-SSM). C-SSM is a cross-channel bi-directional selective scanning module that focuses on extracting cross-channel dependencies.
Like SM block, the input of CM block also consists of two parts: one is the input of the previous block \(Y^{(n-1)}\),   another is the multi-scale input of the current scale \(X_{n}\), the process of the current block can be defined as follows:

\small
\begin{equation}
\begin{aligned}
    X_{in} = X_{n}&+Y^{(n-1)}_c, \\
    X_{mid} = C\text{-}SSM(L&N(X_{in})) + X_{in}, \\
    Y^{(n)}_c = FF\text{-}MoE(L&N(X_{mid})) + X_{mid}
\end{aligned}
\end{equation}

\subsubsection{Channel Selective Scan Module.}
In terms of channel information modeling, existing Mamba architectures primarily scan images from four spatial directions, without considering dependencies across channels. To address this limitation, we extended the 2D-SSM to achieve cross-channel scanning within the Channel Selective Scan module. As illustrated, the Channel-SSM first applies an Global Average Pooling(GAP) operation to the input \( X \) to obtain pooled features. Next, pooled feature is processed through a SiLU activation function and an SSM layer. Finally, a sigmoid activation function is used to generate the channel attention, which is then multiplied with the original input \( X \) to produce the output of the Channel-SSM, which is defined as follows:

\begin{equation}
\begin{aligned}
    X_c = \sigma(SSM(SiLU(&Conv(GAP(X))))),\\
    X_{cout} = X& \odot X_c ,
\end{aligned}
\end{equation}

\begin{figure*}[t]
\centering
\setlength{\tabcolsep}{1pt}
\renewcommand{\arraystretch}{0.7}
\begin{tabular}{cccccccc}
        \textbf{Input} & \textbf{Ucolor}  & \textbf{Ushape} & \textbf{CECF}  & \textbf{HCLR} & \textbf{DM-Water} & \textbf{MambaIR} & \textbf{Ours}
        \\
        \includegraphics[width=0.11\textwidth, height=0.07\textwidth]{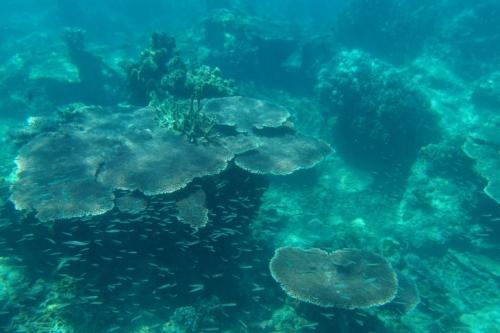} & 
        \includegraphics[width=0.11\textwidth, height=0.07\textwidth]{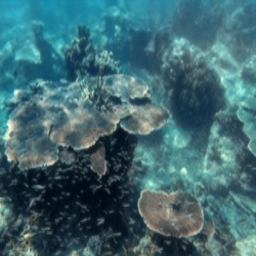} & 
        \includegraphics[width=0.11\textwidth, height=0.07\textwidth]{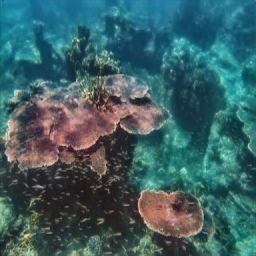} & 
        \includegraphics[width=0.11\textwidth, height=0.07\textwidth]{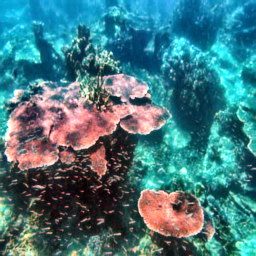} & 
        \includegraphics[width=0.11\textwidth, height=0.07\textwidth]{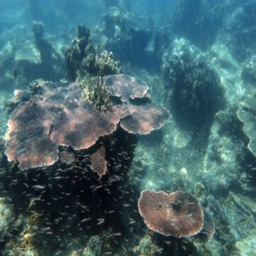} & 
        \includegraphics[width=0.11\textwidth, height=0.07\textwidth]{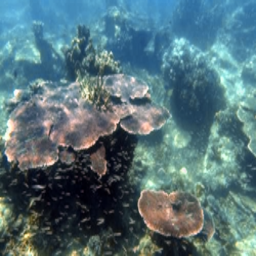}    & 
        \includegraphics[width=0.11\textwidth, height=0.07\textwidth]{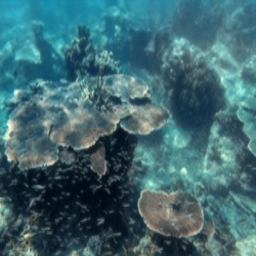} & 
        \includegraphics[width=0.11\textwidth, height=0.07\textwidth]{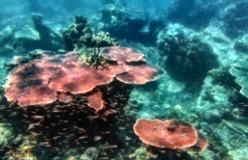} 
        \\
        \includegraphics[width=0.11\textwidth, height=0.07\textwidth]{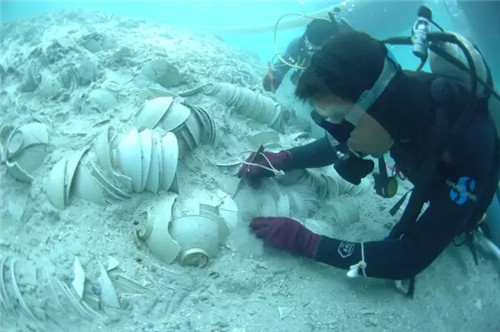} & 
        \includegraphics[width=0.11\textwidth, height=0.07\textwidth]{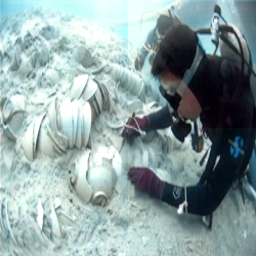} & 
        \includegraphics[width=0.11\textwidth, height=0.07\textwidth]{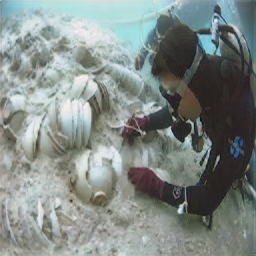} & 
        \includegraphics[width=0.11\textwidth, height=0.07\textwidth]{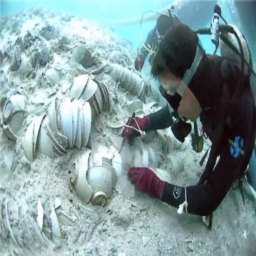} & 
        \includegraphics[width=0.11\textwidth, height=0.07\textwidth]{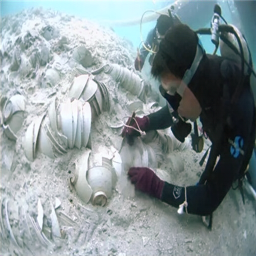} & 
        \includegraphics[width=0.11\textwidth, height=0.07\textwidth]{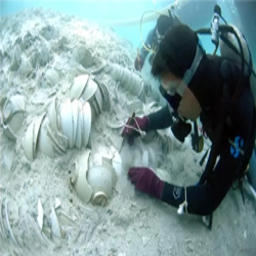}    & 
        \includegraphics[width=0.11\textwidth, height=0.07\textwidth]{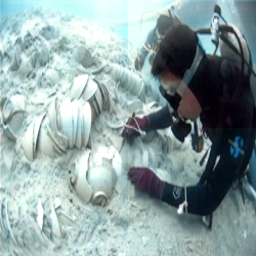} & 
        \includegraphics[width=0.11\textwidth, height=0.07\textwidth]{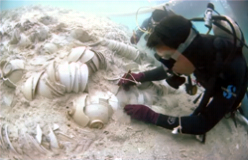} 
        \\
        \includegraphics[width=0.11\textwidth, height=0.07\textwidth]{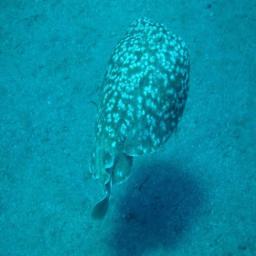} & 
        \includegraphics[width=0.11\textwidth, height=0.07\textwidth]{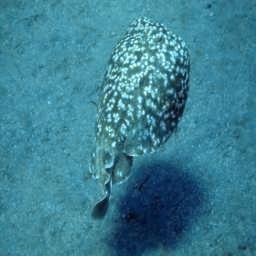} & 
        \includegraphics[width=0.11\textwidth, height=0.07\textwidth]{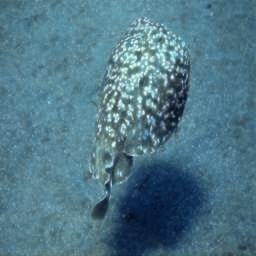} & 
        \includegraphics[width=0.11\textwidth, height=0.07\textwidth]{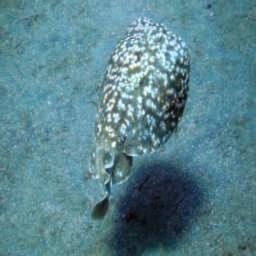} & 
        \includegraphics[width=0.11\textwidth, height=0.07\textwidth]{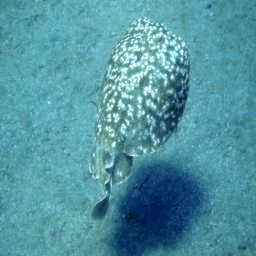} & 
        \includegraphics[width=0.11\textwidth, height=0.07\textwidth]{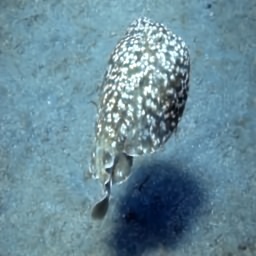}    & 
        \includegraphics[width=0.11\textwidth, height=0.07\textwidth]{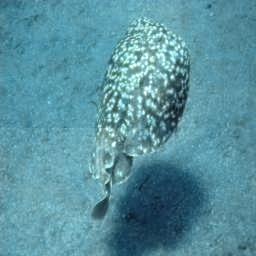} & 
        \includegraphics[width=0.11\textwidth, height=0.07\textwidth]{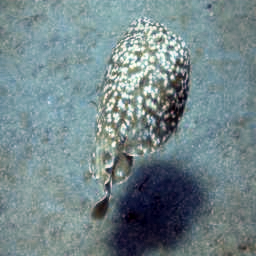} 
        \\
        \includegraphics[width=0.11\textwidth, height=0.07\textwidth]{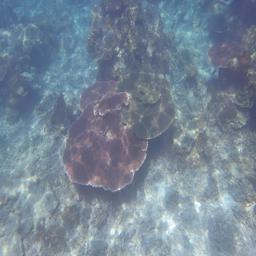} & 
        \includegraphics[width=0.11\textwidth, height=0.07\textwidth]{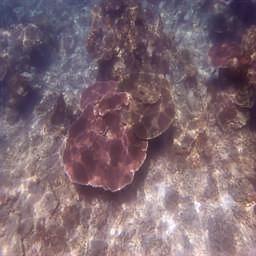} & 
        \includegraphics[width=0.11\textwidth, height=0.07\textwidth]{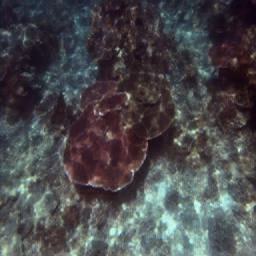} & 
        \includegraphics[width=0.11\textwidth, height=0.07\textwidth]{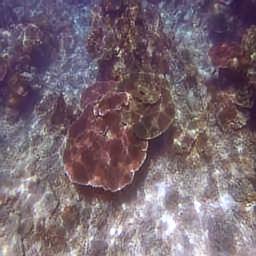} & 
        \includegraphics[width=0.11\textwidth, height=0.07\textwidth]{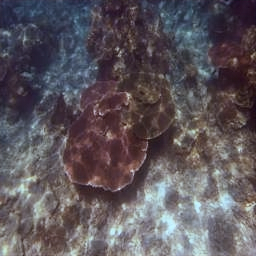} & 
        \includegraphics[width=0.11\textwidth, height=0.07\textwidth]{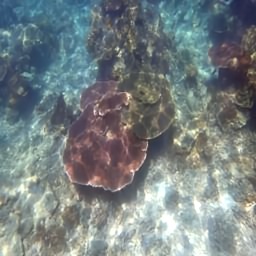}    & 
        \includegraphics[width=0.11\textwidth, height=0.07\textwidth]{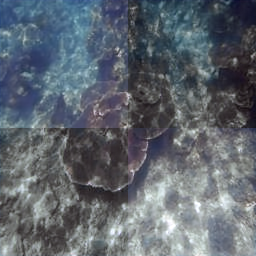} & 
        \includegraphics[width=0.11\textwidth, height=0.07\textwidth]{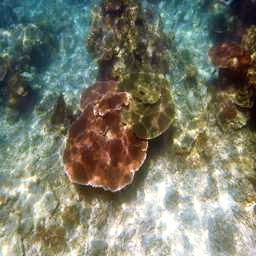} 
        \\
        \includegraphics[width=0.11\textwidth, height=0.07\textwidth]{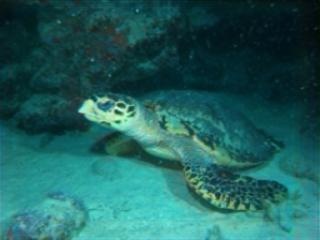} & 
        \includegraphics[width=0.11\textwidth, height=0.07\textwidth]{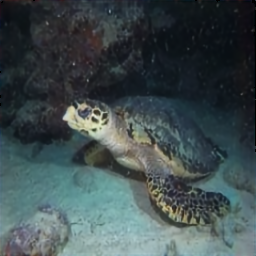} & 
        \includegraphics[width=0.11\textwidth, height=0.07\textwidth]{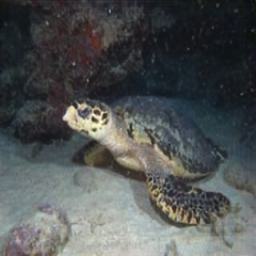} & 
        \includegraphics[width=0.11\textwidth, height=0.07\textwidth]{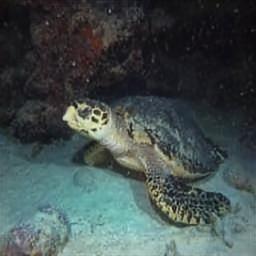} & 
        \includegraphics[width=0.11\textwidth, height=0.07\textwidth]{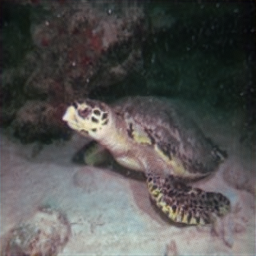} & 
        \includegraphics[width=0.11\textwidth, height=0.07\textwidth]{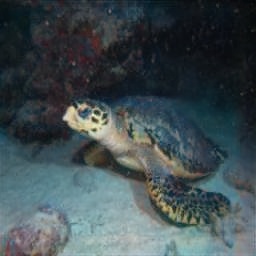}    & 
        \includegraphics[width=0.11\textwidth, height=0.07\textwidth]{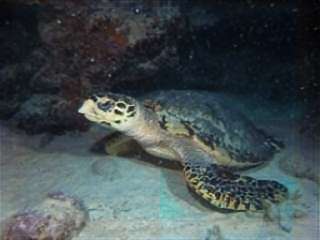} & 
        \includegraphics[width=0.11\textwidth, height=0.07\textwidth]{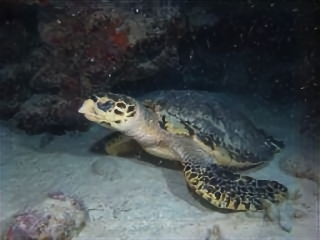} 
        \\
        \includegraphics[width=0.11\textwidth, height=0.07\textwidth]{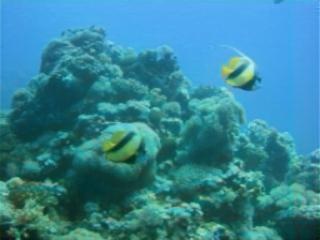} & 
        \includegraphics[width=0.11\textwidth, height=0.07\textwidth]{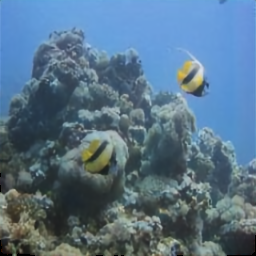} & 
        \includegraphics[width=0.11\textwidth, height=0.07\textwidth]{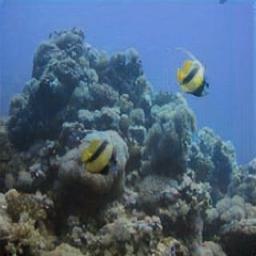} & 
        \includegraphics[width=0.11\textwidth, height=0.07\textwidth]{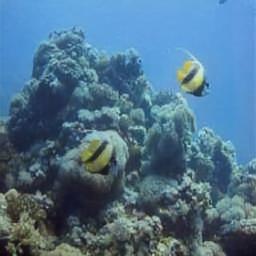} & 
        \includegraphics[width=0.11\textwidth, height=0.07\textwidth]{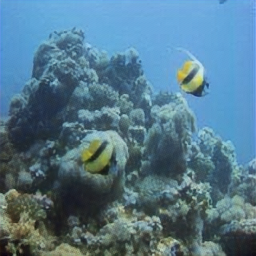} & 
        \includegraphics[width=0.11\textwidth, height=0.07\textwidth]{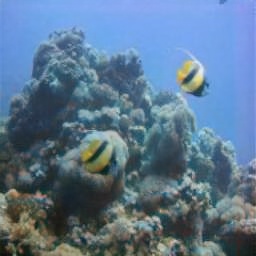}    & 
        \includegraphics[width=0.11\textwidth, height=0.07\textwidth]{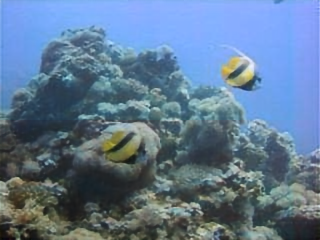} & 
        \includegraphics[width=0.11\textwidth, height=0.07\textwidth]{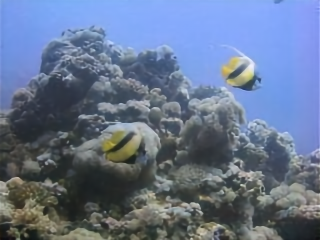} 
        \\
        \includegraphics[width=0.11\textwidth, height=0.07\textwidth]{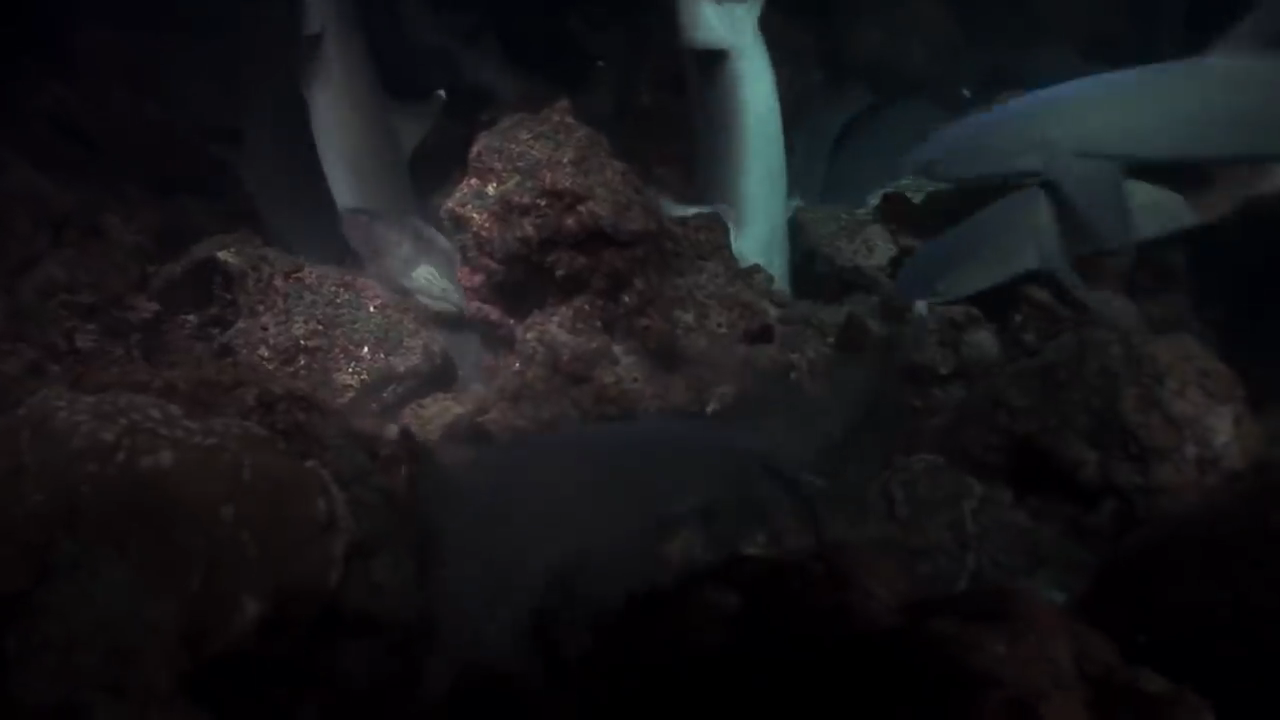} & 
        \includegraphics[width=0.11\textwidth, height=0.07\textwidth]{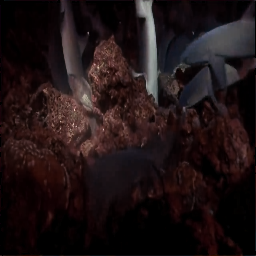} & 
        \includegraphics[width=0.11\textwidth, height=0.07\textwidth]{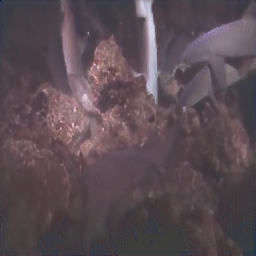} & 
        \includegraphics[width=0.11\textwidth, height=0.07\textwidth]{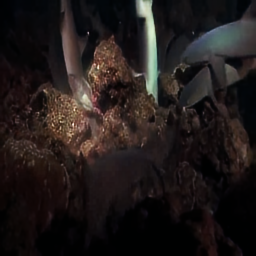} & 
        \includegraphics[width=0.11\textwidth, height=0.07\textwidth]{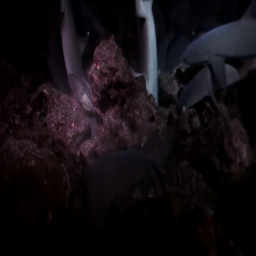} & 
        \includegraphics[width=0.11\textwidth, height=0.07\textwidth]{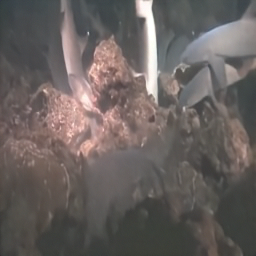}    & 
        \includegraphics[width=0.11\textwidth, height=0.07\textwidth]{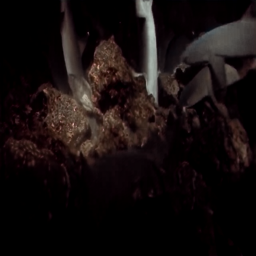} & 
        \includegraphics[width=0.11\textwidth, height=0.07\textwidth]{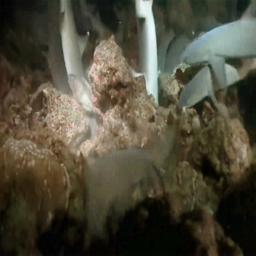} 
        \\
        \includegraphics[width=0.11\textwidth, height=0.07\textwidth]{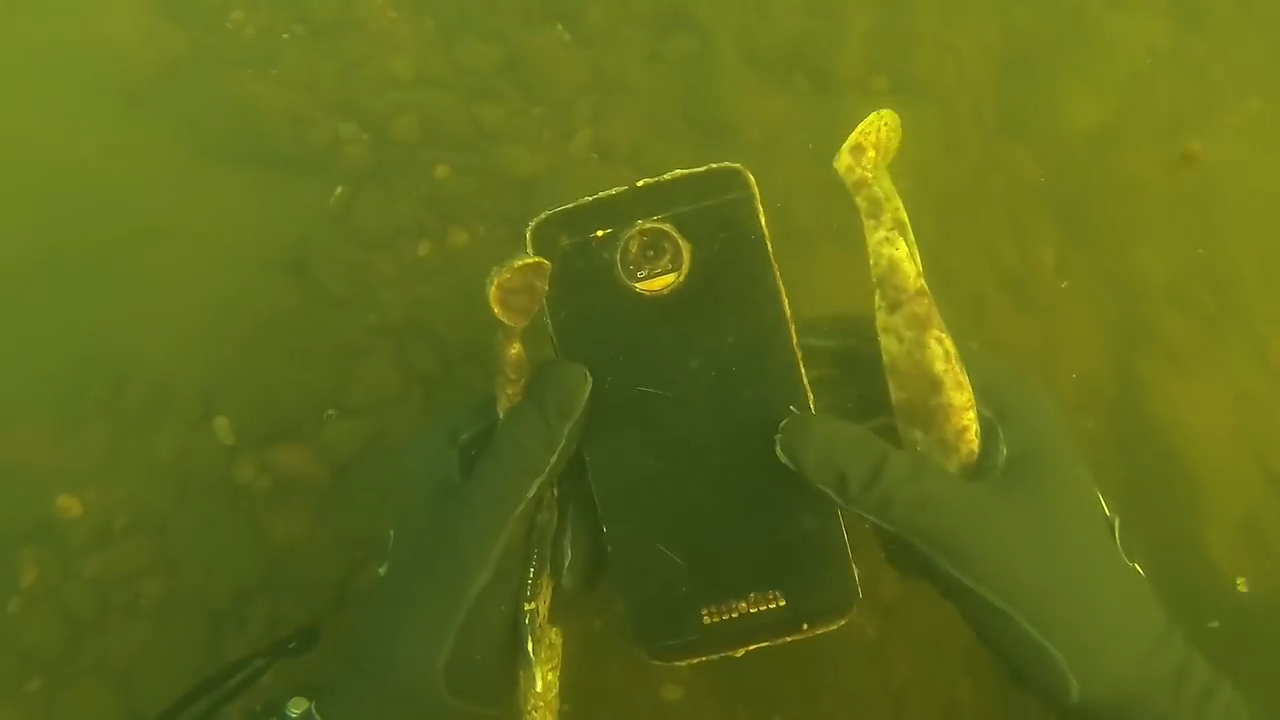} & 
        \includegraphics[width=0.11\textwidth, height=0.07\textwidth]{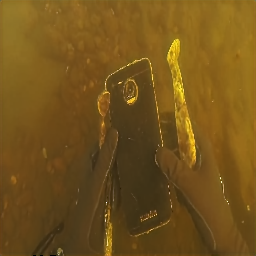} & 
        \includegraphics[width=0.11\textwidth, height=0.07\textwidth]{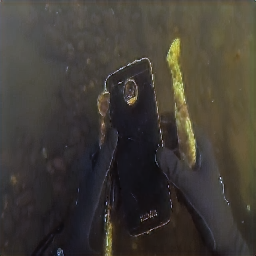} & 
        \includegraphics[width=0.11\textwidth, height=0.07\textwidth]{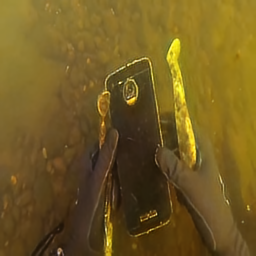} & 
        \includegraphics[width=0.11\textwidth, height=0.07\textwidth]{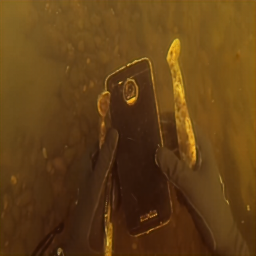} & 
        \includegraphics[width=0.11\textwidth, height=0.07\textwidth]{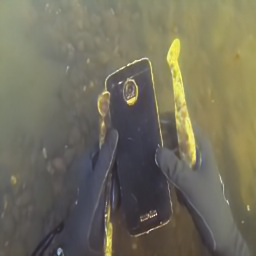}    & 
        \includegraphics[width=0.11\textwidth, height=0.07\textwidth]{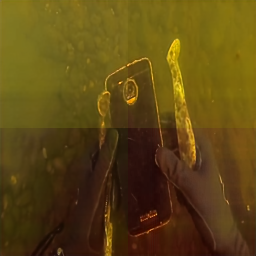} & 
        \includegraphics[width=0.11\textwidth, height=0.07\textwidth]{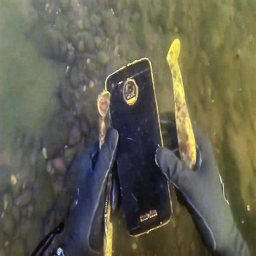} 
        \\
    \end{tabular}
\caption{Comparison of inputs and results from different methods.}
\label{cpp}
\end{figure*}

\subsection{Multi-scale Bi-mutual Promotion Module}
\subsubsection{Multi-scale Mamba MoE}
Compared to FF-MoE, which handles different underwater conditions, the role of Multi-scale Mamba Mixture of Experts (MS-MoE) is to fuse multi-scale features extracted by different Mamba blocks. To address the complexities of varying scale features in different underwater scenarios, MS-MoE uses the Mamba blocks from the current branch as its experts.
Take the spatial Mamba branch as an example. Suppose the outputs of the S-SSMs in the encoder are \(Y_1\), \(Y_2\) and \(Y_3\). Upon receiving these features, MS-MoE first passes them through a convolutional layer to unify their dimensions and concatenates them as \(Y_{cat}\) for subsequent processing. The combined result is then fed into the Multi-scale Mamba Expert module to obtain \(Y_{s}\). The process is as follows:

\begin{equation}
\begin{aligned}
    Y_{cat} = Concate(Conv_1(Y_1), Conv&_1(Y_2), Conv_1(Y_3)), \\
    Y_{s} = \sum_{i=1}^{N} Expert_i(Y_{cat})& \cdot \mathrm{Gate}(Y_{cat}), \\
\end{aligned}
\end{equation}

Here, Expert is S-SSM block. The results of channel branching \(Y_c\) are similar to the above.

\subsubsection{Mutual Promotion Moudle}
To achieve the interaction between spatial and channel information, we introduce the Spatial-Channel Branch MS-MoE Mutual Promotion mechanism. For the outputs from the Spatial Branch MS-MoE (\(Y_s\)) and the Channel Branch MS-MoE (\(Y_c\)), we combine them by performing element-wise addition to obtain the fused feature \(F\). This fused feature is then processed by three different convolutional MoEs, adjusting it to the same dimensions as the Unet skip connections, resulting in \(Z_{i}\) where \(i \in \{1, 2, 3\}\). These features are then added to the respective skip connections.
The process is as follows:

\begin{equation}
\begin{aligned}
    F = Y_s &+ Y_c \\
    Z_{i} = MoE_i&(F) + Skip_i. \
\end{aligned}
\end{equation}

Here, \(\mathrm{MoE}_i\) represents the convolutional MoE used for processing and adjusting, and \(\mathrm{Skip}_i\) represents the skip connections, where \(i \in \{1, 2, 3\}\).

\subsubsection{Cyclic Multi-scale Optimization Strategy}
We denote \( S_1, S_{\frac{1}{2}}, S_{\frac{1}{4}} \) as the output of the Spatial Mamba branch, and \( C_1, C_{\frac{1}{2}}, C_{\frac{1}{4}} \) as the output of the Channel Mamba branch. The corresponding Ground Truths are denoted as \( G_1, G_{\frac{1}{2}}, G_{\frac{1}{4}} \) respectively. The multi-scale loss can be expressed as:

\begin{equation}
\begin{aligned}
\mathcal{L}_{ms} =  \| G_{(i)} -( S_{(i)}+C_{(i)} )\|_1,
\end{aligned}
\end{equation}

where \(i \in \{1,\frac{1}{2},\frac{1}{4} \}  \).
As shown in Algorithm \ref{loss}, we optimize only one of the three losses during each iteration to reduce interference between different scales, rotating through them in subsequent iterations. This approach, referred to as the Cyclic multi-scale optimization strategy, enhances the model’s performance by preventing potential conflicts that could arise from simultaneously optimizing multiple scale losses.

\begin{algorithm}[h]
\caption{Cyclic Multi-scale Optimization strategy}
\begin{algorithmic}[1]
\REQUIRE Current iteration \(k\), Ground Truths \(G_1, G_{\frac{1}{2}}, G_{\frac{1}{4}}\)
\FOR{each training iteration \(k\)}
    \STATE Get the outputs of SM branch: \(S_1, S_{\frac{1}{2}}, S_{\frac{1}{4}}\)
    \STATE Get the outputs of CM branch: \(C_1, C_{\frac{1}{2}}, C_{\frac{1}{4}}\)
    \IF{\(k \bmod 3 == 0\)}
        \STATE \(\mathcal{L}_{ms} = \| G_1 - ( C_{1} +S_1 ) \|_1\)
    \ELSIF{\(k \bmod 3 == 1\)}
        \STATE \(\mathcal{L}_{ms} = \| G_{\frac{1}{2}} - (C_{\frac{1}{2}} + S_{\frac{1}{2}}) \|_1\)
    \ELSIF{\(k \bmod 3 == 2\)}
        \STATE \(\mathcal{L}_{ms} = \| G_{\frac{1}{4}} - (C_{\frac{1}{4}} + S_{\frac{1}{4}}) \|_1\)
    \ENDIF
    \STATE Backpropagate \(\mathcal{L}_{ms}\) and update the parameters
\ENDFOR
\end{algorithmic}
\label{loss}
\end{algorithm}

\begin{table*}[ht]
\setlength{\tabcolsep}{3.5pt}
\renewcommand{\arraystretch}{1.5} 
\centering
\begin{tabular}{c|cccc|cc|cc|cc|cc}
\toprule
& \multicolumn{4}{c|}{\textbf{UIEB}}  &  \multicolumn{2}{c|}{\textbf{LSUI}} &  \multicolumn{2}{c|}{\textbf{EUVP}}&  \multicolumn{2}{c|}{\textbf{U45}} &  \multicolumn{2}{c}{\textbf{Challenge-60}}\\
\multirow{-2}{*}{\Large Methods}& PSNR$\uparrow$ & SSIM$\uparrow$ & LPIPS$\downarrow$ & FID$\downarrow$ 
& PSNR$\uparrow$ & SSIM$\uparrow$
& PSNR$\uparrow$ & SSIM$\uparrow$ 
& UIQM$\uparrow$ &URanker$\uparrow$
& UIQM$\uparrow$ &URanker$\uparrow$
\\
\midrule
Ucolor   & 20.08 & 0.8906 & 0.1205 & 39.63 
         & 24.47 & 0.8989 
         & 25.50 & 0.8561  
         & 3.9913  & 0.9749   
         & 3.6622 & 0.8133
         \\
GUPDM    & 21.95 & 0.9068 & \underline{0.1143} & 29.65  
         & 25.20 & 0.8976    
         & 25.28 & 0.8525 
         & 4.1537  & 1.3139   
         & 3.7598 & 0.8757 
         
         \\
U-shape  & 21.74 & 0.8207 & 0.1260 & 53.89    
         & 26.18 & 0.8823  
         & 25.61 & 0.8642 
         & 4.1271 & 1.1998  
         & 3.8543 & 0.8638
         \\
CECF     & 22.99 & 0.9046 & 0.1193 & 33.84  
         & 26.16 & 0.8656  
         & 28.77 & 0.8864  
         & \textbf{4.2268}  & \underline{1.4464}   
         & 3.9037 & 1.0820
         \\
HCLR-Net & 23.17 & 0.9215 & 0.1146 & 38.11
         & 27.17 & 0.9075   
         & 27.58 & 0.8681   
         & 4.1456  & 1.3648    
         & 3.9018 & 1.0233 
         \\
DM-water & 24.03 & 0.9247 & 0.1346 & 31.02 
         & 28.95 & 0.9083     
         & 29.74 & \underline{0.9297}  
         & 4.0992  & 1.4025     
         & 3.8042 & \underline{1.0947}
         \\
MambaIR  & 24.71 & 0.9399 & 0.1407 & 50.78  
         & 28.05 & 0.9172     
         & 29.90 & 0.9114 
         & \underline{4.2096}  & 1.4278  
         & 3.9166 & 1.0405
         \\
\midrule
Ours-light& \underline{25.53} & \underline{0.9519} & 0.1158 & \underline{27.42}    
         & \underline{29.25}  & \underline{0.9154} 
         & \underline{31.54}  & 0.9084
         & 4.1253  & 1.4436
         & \underline{3.9192}  & 1.0893
         \\
Ours     & \textbf{26.78} & \textbf{0.9597} & \textbf{0.1133} & \textbf{25.30}    
         & \textbf{30.47} &\textbf{ 0.9245}    
         & \textbf{32.06} & \textbf{0.9194}
         & 4.1594         & \textbf{1.4539} 
         & \textbf{3.9231}  & \textbf{1.1100}
         \\
\bottomrule
\end{tabular}
\caption{Quantitative comparison on the UIEBD, LSUI, EUVP, U45 and Challenge-60 datasets, The best results are highlighted in bold and the second best results are underlined.}
\label{cpm}
\end{table*}

\setlength{\tabcolsep}{3pt}{
\renewcommand{\arraystretch}{1.2} 
\begin{table}[h]
\centering
\begin{tabular}{c|c|c}
\toprule
Methods  &Params (M)  &FLOPs (G) \\
\midrule
Ucolor     & 157.4  & 443   \\
U-shape    & 65.6   & \textbf{66}   \\
MambaIR    & \textbf{16.7}   & 654    \\
\midrule
Ours-light & \underline{25.7}   & \underline{156}     \\
Ours       & 57.8   & 300   \\
\bottomrule
\end{tabular}
\caption{Comparison of Params and FLOPs}
\label{time}
\end{table}
}

\begin{table}[h]
\centering
\setlength{\tabcolsep}{3pt}
\renewcommand{\arraystretch}{1.2} 
\begin{tabular}{c|ccc|cc}
\toprule
& \multicolumn{3}{c|}{\textbf{Settings}} & \multicolumn{2}{c}{\textbf{Metrics}}\\
\multirow{-2}{*}{\Large \textbf{Methods}} & SMB & CMB & MP &PSNR$\uparrow$ & SSIM$\uparrow$ \\
\midrule
O-Mamba-C                &\ding{55} &\ding{51}  &\ding{51} & 28.59 & 0.9139 \\
O-Mamba-S                &\ding{51} &\ding{55}  &\ding{51} & 29.31 & 0.9187 \\
O-Mamba w/o MP           &\ding{51} &\ding{51}  &\ding{55} & 29.82 & 0.9236 \\
\midrule
O-Mamba                  &\ding{51} &\ding{51}  &\ding{51} &\textbf{30.47} & \textbf{0.9245} \\

\bottomrule
\end{tabular}
\caption{Ablation studies with Framework. O-Mamba-S refers to use single Spatial Mamba Branch, O-Mamba-C refers to use single Channel Mamba Branch. }
\label{as}
\end{table}

\begin{table}[h]
\centering
\setlength{\tabcolsep}{6pt}
\renewcommand{\arraystretch}{1.3} 
\begin{tabular}{c|cc}
\toprule
& \multicolumn{2}{c}{\textbf{Metrics}}\\
\multirow{-2}{*}{\Large \textbf{Methods}}  & PSNR$\uparrow$ & SSIM$\uparrow$  \\
\midrule
O-Mamba w/o MS-Input      & 29.26 & 0.9039  \\
O-Mamba w/o FF-MoE        & 29.13 & 0.9226 \\
O-Mamba w/o MS-MoE        & 28.82 & 0.9173 \\
O-Mamba w/o CMS-OS        & 29.98 & 0.9151 \\
\midrule
O-Mamba             &\textbf{30.47} & \textbf{0.9245} \\
\bottomrule
\end{tabular}

\caption{Ablation studies with Modules. MS-Input refers multi-scale inputs, CMS-OS refers cyclic multi-scale optimization strategy.}
\label{as2}
\end{table}

\section{EXPERIMENTS}
\subsection{Experimental Settings}
\subsubsection{Implementation details.}
We implemented O-Mamba using PyTorch on an NVIDIA GeForce RTX 3090 GPU, utilizing an Adam optimizer (\( \beta_1 = 0.9, \beta_2 = 0.999\)) with an initial learning rate of \(2 \times 10^{-4}\), halving it at specific training milestones. The training consisted of 500K iterations with a batch size of 4 and an input patch size of \(128 \times 128\).

\subsubsection{Datasets.}
We utilize the UIEBD dataset \cite{8917818} and the LSUI dataset \cite{peng2023u} for training and evaluating our model. The UIEBD dataset comprises 890 underwater images with corresponding labels. We use 800 images for training, and the remaining 90 images for testing. The LSUI dataset is randomly partitioned into 4604 images for training and 400 images for testing. In addition, to verify the generalization of O-Mamba, we utilize the reference benchmark EUVP \cite{islam2020fast}, which contains 515 paired test samples, along with the non-reference benchmarks U45 \cite{li2019fusionadversarialunderwaterimage} and Challenge-60 \cite{8917818}, which contain 45 and 60 underwater images for testing, respectively.
\subsubsection{Comparison Methods.}
We perform a comparative analysis of O-Mamba against 6 state-of-the-art (SOTA) UIE methods, namely Ucolor \cite{Ucolor},GUPDM \cite{mu2023generalized}, DM-water \cite{tang2023underwater}, U-shape \cite{peng2023u}, 
CECF \cite{cong2024underwater}, HCLR-Net \cite{zhou2024hclr} 
and a SOTA Mamba method namely MambaIR \cite{guo2024mambairsimplebaselineimage}. 
In addition to O-Mamba, we also present a O-Mamba-light by optimizing modules. Mamba-light delivers state-of-the-art (SOTA) performance with half the parameters and FLOPS cost compared to the original Mamba.
To ensure a fair and rigorous comparison, we utilize the provided source codes from the respective authors and adhere strictly to the identical experimental settings across all evaluations.

\subsubsection{Evaluation Metrics.}
We use well-established full-reference metrics: PSNR and SSIM, which offer quantitative comparisons at both pixel and structural levels. Higher values indicate better image quality. Additionally, we evaluate LPIPS \cite{zhang2018unreasonable} and FID \cite{heusel2017gans}; LPIPS assesses perceptual similarity, while FID measures the distance between real and generated image distributions, with lower scores indicating better performance. Due to space constraints, LPIPS and FID were tested only on the UIEB dataset. For non-reference benchmarks U45 and UIEB-challenge, we employ UIQM \cite{panetta2015human} and Uranker \cite{guo2023underwater} for evaluation.

\subsection{Results and Comparisons.}
Table \ref{cpm} presents the quantitative results of O-Mamba compared to seven state-of-the-art methods on the LSUI, UIEB, EUVP, U45, and Challenge-60 datasets. The results show that O-Mamba consistently outperforms other methods across most metrics, achieving SOTA performance and demonstrating the effectiveness and robustness of our approach in UIE tasks.
Figure \ref{cpp} provides a visual comparison, showcasing O-Mamba's superiority in light, detail, and color enhancement across representative underwater scenes from the five datasets. This validates the effectiveness of our dual-branch approach for processing cross-channel information and highlights its potential for real-world applications.
We selected three representative methods based on convolutional neural networks, Transformers, and Mamba to compare parameter counts and computational complexity as shown in table \ref{time}. The results indicate that while O-Mamba has a relatively large parameter count due to its dual branches and MoE, it remains lower than Ucolor (convolution-based) and Ushape (Transformer-based). Despite having three times the parameters of MambaIR, O-Mamba's computational complexity is more less than MambaIR. 

\subsection{Ablation Study}
In order to validate the effectiveness of O-Mamba, we design two ablation studies: one focusing on the overall framework, with the results presented in Table \ref{as}, and the other examining each individual module, with the results presented in Table \ref{as2}.

\subsubsection{Ablation study with overall framework.}
To evaluate the overall architecture of O-Mamba, we conducted ablation experiments on each branch and their interactions using the LSUI dataset. The results are shown in Table \ref{as}. When only the spatial Mamba branch (O-Mamba-S) or the channel Mamba branch (O-Mamba-C) is used, O-Mamba degrades into a Mamba-based Unet with MS-MoE. The data shows that with only the spatial branch, the PSNR is 29; with only the channel branch, the PSNR is 28; and with the dual-branch structure, the PSNR reaches 30. Therefore, we can conclude that the spatial Mamba branch and the channel Mamba branch model the underwater image enhancement process from different dimensions, with the spatial Mamba branch playing the leading role and the channel Mamba branch supplementing it from the channel dimension, creating effective complementarity. When the MP module is removed, the PSNR is 29.82; with the MP module, the PSNR increases to 30, indicating that the MP module effectively facilitates inter-branch information interaction. Without the MP module, the two branches simply combine their results through basic addition, making it challenging to fully leverage the complex spatial and channel information.

\subsubsection{Ablation study with each module.}
To evaluate the effectiveness of each component within O-Mamba, we conducted ablation experiments on the LSUI dataset. As shown in Table \ref{as2}, the removal of any module resulted in varying degrees of performance degradation, demonstrating the importance of each module for the UIE task. Furthermore, the table highlights that when MS-MoE was removed, the model suffered significant performance loss, indicating that integrating multi-scale information for dual-branch interaction is crucial. This multi-scale cross-branch interaction effectively captures complementary information across dimensions, improving the model's overall generalization capability.

\section{Conclusion}
This paper presents a novel underwater image enhancement framework called O-Mamba, an O-shaped dual-branch Mamba network that extracts global dependencies in underwater images from both spatial and cross-channel dimensions. The dual branches interact through MSBMP Module, enabling O-Mamba to comprehensively model underwater images. Experimental results show that O-Mamba achieves state-of-the-art performance across multiple datasets. Ablation studies further validate the effectiveness of the O-shaped structure in the task of underwater image enhancement.

\bibliography{Omamba}

\begin{thebibliography}{62}
\providecommand{\natexlab}[1]{#1}

\bibitem[{{Abdul Ghani} and {Mat Isa}(2017)}]{ABDULGHANI2017181}
{Abdul Ghani}, A.~S.; and {Mat Isa}, N.~A. 2017.
\newblock Automatic system for improving underwater image contrast and color through recursive adaptive histogram modification.
\newblock \emph{Computers and Electronics in Agriculture}, 141: 181--195.

\bibitem[{Akkaynak and Treibitz(2018)}]{akkaynak2018revised}
Akkaynak, D.; and Treibitz, T. 2018.
\newblock A revised underwater image formation model.
\newblock In \emph{Proceedings of the IEEE conference on computer vision and pattern recognition}, 6723--6732.

\bibitem[{Anwar and Li(2020)}]{anwar2020diving}
Anwar, S.; and Li, C. 2020.
\newblock Diving deeper into underwater image enhancement: A survey.
\newblock \emph{Signal Processing: Image Communication}, 89: 115978.

\bibitem[{Berman et~al.(2020)Berman, Levy, Avidan, and Treibitz}]{berman2020underwater}
Berman, D.; Levy, D.; Avidan, S.; and Treibitz, T. 2020.
\newblock Underwater single image color restoration using haze-lines and a new quantitative dataset.
\newblock \emph{IEEE transactions on pattern analysis and machine intelligence}, 43(8): 2822--2837.

\bibitem[{Cheng, Wang, and Sun(2024)}]{cheng2024activatingwiderareasimage}
Cheng, C.; Wang, H.; and Sun, H. 2024.
\newblock Activating Wider Areas in Image Super-Resolution.
\newblock arXiv:2403.08330.

\bibitem[{Cong, Gui, and Hou(2024)}]{cong2024underwater}
Cong, X.; Gui, J.; and Hou, J. 2024.
\newblock Underwater Organism Color Fine-Tuning via Decomposition and Guidance.
\newblock \emph{Proceedings of the AAAI Conference on Artificial Intelligence}, 38(2): 1389--1398.

\bibitem[{Cong et~al.(2024)Cong, Zhao, Gui, Hou, and Tao}]{cong2024comprehensivesurveyunderwaterimage}
Cong, X.; Zhao, Y.; Gui, J.; Hou, J.; and Tao, D. 2024.
\newblock A Comprehensive Survey on Underwater Image Enhancement Based on Deep Learning.
\newblock arXiv:2405.19684.

\bibitem[{de~Langis and Sattar(2020)}]{9197308}
de~Langis, K.; and Sattar, J. 2020.
\newblock Realtime Multi-Diver Tracking and Re-identification for Underwater Human-Robot Collaboration.
\newblock In \emph{2020 IEEE International Conference on Robotics and Automation (ICRA)}, 11140--11146.

\bibitem[{Drews et~al.(2013)Drews, Nascimento, Moraes, Botelho, and Campos}]{drews2013transmission}
Drews, P.; Nascimento, E.; Moraes, F.; Botelho, S.; and Campos, M. 2013.
\newblock Transmission estimation in underwater single images.
\newblock In \emph{Proceedings of the IEEE international conference on computer vision workshops}, 825--830.

\bibitem[{Elfwing, Uchibe, and Doya(2018)}]{ELFWING20183}
Elfwing, S.; Uchibe, E.; and Doya, K. 2018.
\newblock Sigmoid-weighted linear units for neural network function approximation in reinforcement learning.
\newblock \emph{Neural Networks}, 107: 3--11.
\newblock Special issue on deep reinforcement learning.

\bibitem[{Fu et~al.(2023)Fu, Dao, Saab, Thomas, Rudra, and Ré}]{fu2023hungryhungryhipposlanguage}
Fu, D.~Y.; Dao, T.; Saab, K.~K.; Thomas, A.~W.; Rudra, A.; and Ré, C. 2023.
\newblock Hungry Hungry Hippos: Towards Language Modeling with State Space Models.
\newblock arXiv:2212.14052.

\bibitem[{Gao et~al.(2019)Gao, Zhang, Zhao, Zhang, and Li}]{gao2019underwater}
Gao, S.-B.; Zhang, M.; Zhao, Q.; Zhang, X.-S.; and Li, Y.-J. 2019.
\newblock Underwater image enhancement using adaptive retinal mechanisms.
\newblock \emph{IEEE Transactions on Image Processing}, 28(11): 5580--5595.

\bibitem[{Gu and Dao(2023)}]{gu2023mamba}
Gu, A.; and Dao, T. 2023.
\newblock Mamba: Linear-time sequence modeling with selective state spaces.
\newblock \emph{arXiv preprint arXiv:2312.00752}.

\bibitem[{Gu and Dao(2024)}]{gu2024mambalineartimesequencemodeling}
Gu, A.; and Dao, T. 2024.
\newblock Mamba: Linear-Time Sequence Modeling with Selective State Spaces.
\newblock arXiv:2312.00752.

\bibitem[{Gu, Goel, and Ré(2022)}]{gu2022efficientlymodelinglongsequences}
Gu, A.; Goel, K.; and Ré, C. 2022.
\newblock Efficiently Modeling Long Sequences with Structured State Spaces.
\newblock arXiv:2111.00396.

\bibitem[{Gu et~al.(2021)Gu, Johnson, Goel, Saab, Dao, Rudra, and R{\'e}}]{gu2021combining}
Gu, A.; Johnson, I.; Goel, K.; Saab, K.; Dao, T.; Rudra, A.; and R{\'e}, C. 2021.
\newblock Combining recurrent, convolutional, and continuous-time models with linear state space layers.
\newblock \emph{Advances in neural information processing systems}, 34: 572--585.

\bibitem[{Guo et~al.(2023)Guo, Wu, Jin, Han, Zhang, Chai, and Li}]{guo2023underwater}
Guo, C.; Wu, R.; Jin, X.; Han, L.; Zhang, W.; Chai, Z.; and Li, C. 2023.
\newblock Underwater ranker: Learn which is better and how to be better.
\newblock In \emph{Proceedings of the AAAI conference on artificial intelligence}, volume~37, 702--709.

\bibitem[{Guo et~al.(2024)Guo, Li, Dai, Ouyang, Ren, and Xia}]{guo2024mambairsimplebaselineimage}
Guo, H.; Li, J.; Dai, T.; Ouyang, Z.; Ren, X.; and Xia, S.-T. 2024.
\newblock MambaIR: A Simple Baseline for Image Restoration with State-Space Model.
\newblock arXiv:2402.15648.

\bibitem[{He et~al.(2016)He, Zhang, Ren, and Sun}]{He_2016_CVPR}
He, K.; Zhang, X.; Ren, S.; and Sun, J. 2016.
\newblock Deep Residual Learning for Image Recognition.
\newblock In \emph{Proceedings of the IEEE Conference on Computer Vision and Pattern Recognition (CVPR)}.

\bibitem[{He et~al.(2024)He, Yan, Li, Xie, Zhang, and Zhou}]{he2024frequency}
He, X.; Yan, K.; Li, R.; Xie, C.; Zhang, J.; and Zhou, M. 2024.
\newblock Frequency-Adaptive Pan-Sharpening with Mixture of Experts.
\newblock In \emph{Proceedings of the AAAI Conference on Artificial Intelligence}, volume~38, 2121--2129.

\bibitem[{Heusel et~al.(2017)Heusel, Ramsauer, Unterthiner, Nessler, and Hochreiter}]{heusel2017gans}
Heusel, M.; Ramsauer, H.; Unterthiner, T.; Nessler, B.; and Hochreiter, S. 2017.
\newblock Gans trained by a two time-scale update rule converge to a local nash equilibrium.
\newblock \emph{Advances in neural information processing systems}, 30.

\bibitem[{Horn(1990)}]{horn1990hadamard}
Horn, R.~A. 1990.
\newblock The hadamard product.
\newblock In \emph{Proc. Symp. Appl. Math}, volume~40, 87--169.

\bibitem[{Hou et~al.(2023)Hou, Li, Zhuang, Li, Sun, and Li}]{hou2023non}
Hou, G.; Li, N.; Zhuang, P.; Li, K.; Sun, H.; and Li, C. 2023.
\newblock Non-uniform illumination underwater image restoration via illumination channel sparsity prior.
\newblock \emph{IEEE Transactions on Circuits and Systems for Video Technology}.

\bibitem[{Huang et~al.(2022)Huang, Li, Hua, and Fan}]{huang2022underwater}
Huang, Z.; Li, J.; Hua, Z.; and Fan, L. 2022.
\newblock Underwater image enhancement via adaptive group attention-based multiscale cascade transformer.
\newblock \emph{IEEE Transactions on Instrumentation and Measurement}, 71: 1--18.

\bibitem[{Islam, Xia, and Sattar(2020)}]{islam2020fast}
Islam, M.~J.; Xia, Y.; and Sattar, J. 2020.
\newblock Fast underwater image enhancement for improved visual perception.
\newblock \emph{IEEE Robotics and Automation Letters}, 5(2): 3227--3234.

\bibitem[{Kalman(1960)}]{kalman1960new}
Kalman, R.~E. 1960.
\newblock A new approach to linear filtering and prediction problems.

\bibitem[{Li et~al.(2021)Li, Anwar, Hou, Cong, Guo, and Ren}]{Ucolor}
Li, C.; Anwar, S.; Hou, J.; Cong, R.; Guo, C.; and Ren, W. 2021.
\newblock Underwater Image Enhancement via Medium Transmission-Guided Multi-Color Space Embedding.
\newblock \emph{IEEE Transactions on Image Processing}, 30.

\bibitem[{Li, Anwar, and Porikli(2020)}]{li2020underwater}
Li, C.; Anwar, S.; and Porikli, F. 2020.
\newblock Underwater scene prior inspired deep underwater image and video enhancement.
\newblock \emph{Pattern Recognition}, 98: 107038.

\bibitem[{Li et~al.(2020)Li, Guo, Ren, Cong, Hou, Kwong, and Tao}]{8917818}
Li, C.; Guo, C.; Ren, W.; Cong, R.; Hou, J.; Kwong, S.; and Tao, D. 2020.
\newblock An Underwater Image Enhancement Benchmark Dataset and Beyond.
\newblock \emph{IEEE Transactions on Image Processing}, 29: 4376--4389.

\bibitem[{Li, Li, and Wang(2019)}]{li2019fusionadversarialunderwaterimage}
Li, H.; Li, J.; and Wang, W. 2019.
\newblock A Fusion Adversarial Underwater Image Enhancement Network with a Public Test Dataset.
\newblock arXiv:1906.06819.

\bibitem[{Liang et~al.(2021)Liang, Cao, Sun, Zhang, Van~Gool, and Timofte}]{liang2021swinir}
Liang, J.; Cao, J.; Sun, G.; Zhang, K.; Van~Gool, L.; and Timofte, R. 2021.
\newblock Swinir: Image restoration using swin transformer.
\newblock In \emph{Proceedings of the IEEE/CVF international conference on computer vision}, 1833--1844.

\bibitem[{Liu et~al.(2020)Liu, Tang, Tharmarasa, Kirubarajan, Jassemi, and Hall{\'e}}]{liu2020underwater}
Liu, B.; Tang, X.; Tharmarasa, R.; Kirubarajan, T.; Jassemi, R.; and Hall{\'e}, S. 2020.
\newblock Underwater target tracking in uncertain multipath ocean environments.
\newblock \emph{IEEE Transactions on Aerospace and Electronic Systems}, 56(6): 4899--4915.

\bibitem[{Liu et~al.(2024)Liu, Tian, Zhao, Yu, Xie, Wang, Ye, and Liu}]{liu2024vmambavisualstatespace}
Liu, Y.; Tian, Y.; Zhao, Y.; Yu, H.; Xie, L.; Wang, Y.; Ye, Q.; and Liu, Y. 2024.
\newblock VMamba: Visual State Space Model.
\newblock arXiv:2401.10166.

\bibitem[{Liu et~al.(2022)Liu, Zhuang, Jia, Wu, Xu, and Liu}]{liu2022novel}
Liu, Z.; Zhuang, Y.; Jia, P.; Wu, C.; Xu, H.; and Liu, Z. 2022.
\newblock A novel underwater image enhancement algorithm and an improved underwater biological detection pipeline.
\newblock \emph{Journal of Marine Science and Engineering}, 10(9): 1204.

\bibitem[{Lu, Liu, and Kong(2023)}]{Lu_2023_ICCV}
Lu, S.; Liu, Y.; and Kong, A. W.-K. 2023.
\newblock TF-ICON: Diffusion-Based Training-Free Cross-Domain Image Composition.
\newblock In \emph{Proceedings of the IEEE/CVF International Conference on Computer Vision (ICCV)}, 2294--2305.

\bibitem[{Lu et~al.(2024)Lu, Wang, Li, Liu, and Kong}]{Lu_2024_CVPR}
Lu, S.; Wang, Z.; Li, L.; Liu, Y.; and Kong, A. W.-K. 2024.
\newblock MACE: Mass Concept Erasure in Diffusion Models.
\newblock In \emph{Proceedings of the IEEE/CVF Conference on Computer Vision and Pattern Recognition (CVPR)}, 6430--6440.

\bibitem[{McMahon and Plaku(2021)}]{9359465}
McMahon, J.; and Plaku, E. 2021.
\newblock Autonomous Data Collection With Timed Communication Constraints for Unmanned Underwater Vehicles.
\newblock \emph{IEEE Robotics and Automation Letters}, 6(2): 1832--1839.

\bibitem[{Mehta et~al.(2022)Mehta, Gupta, Cutkosky, and Neyshabur}]{mehta2022longrangelanguagemodeling}
Mehta, H.; Gupta, A.; Cutkosky, A.; and Neyshabur, B. 2022.
\newblock Long Range Language Modeling via Gated State Spaces.
\newblock arXiv:2206.13947.

\bibitem[{Mei et~al.(2022)Mei, Ye, Zhang, Liu, Wang, Hou, and Wang}]{mei2022uir}
Mei, X.; Ye, X.; Zhang, X.; Liu, Y.; Wang, J.; Hou, J.; and Wang, X. 2022.
\newblock Uir-net: a simple and effective baseline for underwater image restoration and enhancement.
\newblock \emph{Remote Sensing}, 15(1): 39.

\bibitem[{Mu et~al.(2023)Mu, Xu, Liu, Wang, Chan, and Bai}]{mu2023generalized}
Mu, P.; Xu, H.; Liu, Z.; Wang, Z.; Chan, S.; and Bai, C. 2023.
\newblock A generalized physical-knowledge-guided dynamic model for underwater image enhancement.
\newblock In \emph{Proceedings of the 31st ACM international conference on multimedia}, 7111--7120.

\bibitem[{Naik, Swarnakar, and Mittal(2021)}]{Naik_Swarnakar_Mittal_2021}
Naik, A.; Swarnakar, A.; and Mittal, K. 2021.
\newblock Shallow-UWnet: Compressed Model for Underwater Image Enhancement (Student Abstract).
\newblock \emph{Proceedings of the AAAI Conference on Artificial Intelligence}, 35(18): 15853--15854.

\bibitem[{Nguyen et~al.(2022)Nguyen, Goel, Gu, Downs, Shah, Dao, Baccus, and R{\'e}}]{nguyen2022s4nd}
Nguyen, E.; Goel, K.; Gu, A.; Downs, G.; Shah, P.; Dao, T.; Baccus, S.; and R{\'e}, C. 2022.
\newblock S4nd: Modeling images and videos as multidimensional signals with state spaces.
\newblock \emph{Advances in neural information processing systems}, 35: 2846--2861.

\bibitem[{Panetta, Gao, and Agaian(2015)}]{panetta2015human}
Panetta, K.; Gao, C.; and Agaian, S. 2015.
\newblock Human-visual-system-inspired underwater image quality measures.
\newblock \emph{IEEE Journal of Oceanic Engineering}, 41(3): 541--551.

\bibitem[{Peng, Zhu, and Bian(2023)}]{peng2023u}
Peng, L.; Zhu, C.; and Bian, L. 2023.
\newblock U-shape transformer for underwater image enhancement.
\newblock \emph{IEEE Transactions on Image Processing}, 32: 3066--3079.

\bibitem[{Peng, Cao, and Cosman(2018)}]{peng2018generalization}
Peng, Y.-T.; Cao, K.; and Cosman, P.~C. 2018.
\newblock Generalization of the dark channel prior for single image restoration.
\newblock \emph{IEEE Transactions on Image Processing}, 27(6): 2856--2868.

\bibitem[{Peng and Cosman(2017)}]{peng2017underwater}
Peng, Y.-T.; and Cosman, P.~C. 2017.
\newblock Underwater image restoration based on image blurriness and light absorption.
\newblock \emph{IEEE transactions on image processing}, 26(4): 1579--1594.

\bibitem[{Pióro et~al.(2024)Pióro, Ciebiera, Król, Ludziejewski, Krutul, Krajewski, Antoniak, Miłoś, Cygan, and Jaszczur}]{pióro2024moemambaefficientselectivestate}
Pióro, M.; Ciebiera, K.; Król, K.; Ludziejewski, J.; Krutul, M.; Krajewski, J.; Antoniak, S.; Miłoś, P.; Cygan, M.; and Jaszczur, S. 2024.
\newblock MoE-Mamba: Efficient Selective State Space Models with Mixture of Experts.
\newblock arXiv:2401.04081.

\bibitem[{Riquelme et~al.(2021)Riquelme, Puigcerver, Mustafa, Neumann, Jenatton, Susano~Pinto, Keysers, and Houlsby}]{riquelme2021scaling}
Riquelme, C.; Puigcerver, J.; Mustafa, B.; Neumann, M.; Jenatton, R.; Susano~Pinto, A.; Keysers, D.; and Houlsby, N. 2021.
\newblock Scaling vision with sparse mixture of experts.
\newblock \emph{Advances in Neural Information Processing Systems}, 34: 8583--8595.

\bibitem[{Smith, Warrington, and Linderman(2023)}]{smith2023simplifiedstatespacelayers}
Smith, J. T.~H.; Warrington, A.; and Linderman, S.~W. 2023.
\newblock Simplified State Space Layers for Sequence Modeling.
\newblock arXiv:2208.04933.

\bibitem[{Tang, Kawasaki, and Iwaguchi(2023)}]{tang2023underwater}
Tang, Y.; Kawasaki, H.; and Iwaguchi, T. 2023.
\newblock Underwater image enhancement by transformer-based diffusion model with non-uniform sampling for skip strategy.
\newblock In \emph{Proceedings of the 31st ACM International Conference on Multimedia}, 5419--5427.

\bibitem[{Wang et~al.(2023)Wang, Lan, Su, and Chen}]{10.1007/978-981-99-7549-5_1}
Wang, M.; Lan, F.; Su, Z.; and Chen, W. 2023.
\newblock Underwater Image Enhancement and Restoration Techniques: A Comprehensive Review, Challenges, and Future Trends.
\newblock In Yongtian, W.; and Lifang, W., eds., \emph{Image and Graphics Technologies and Applications}, 3--18. Singapore: Springer Nature Singapore.
\newblock ISBN 978-981-99-7549-5.

\bibitem[{Wang et~al.(2021)Wang, Cao, Zhang, Wu, and Zha}]{wang2021leveraging}
Wang, Y.; Cao, Y.; Zhang, J.; Wu, F.; and Zha, Z.-J. 2021.
\newblock Leveraging deep statistics for underwater image enhancement.
\newblock \emph{ACM Transactions on Multimedia Computing, Communications, and Applications (TOMM)}, 17(3s): 1--20.

\bibitem[{Wen et~al.(2024)Wen, Cui, Yang, Zhao, Zhai, Gao, Dou, and Chen}]{wen2024waterformer}
Wen, J.; Cui, J.; Yang, G.; Zhao, B.; Zhai, Y.; Gao, Z.; Dou, L.; and Chen, B.~M. 2024.
\newblock WaterFormer: A Global--Local Transformer for Underwater Image Enhancement With Environment Adaptor.
\newblock \emph{IEEE Robotics \& Automation Magazine}.

\bibitem[{Zhang et~al.(2018)Zhang, Isola, Efros, Shechtman, and Wang}]{zhang2018unreasonable}
Zhang, R.; Isola, P.; Efros, A.~A.; Shechtman, E.; and Wang, O. 2018.
\newblock The unreasonable effectiveness of deep features as a perceptual metric.
\newblock In \emph{Proceedings of the IEEE conference on computer vision and pattern recognition}, 586--595.

\bibitem[{Zhao et~al.(2024{\natexlab{a}})Zhao, Cai, Dong, and Hu}]{zhao2024wavelet}
Zhao, C.; Cai, W.; Dong, C.; and Hu, C. 2024{\natexlab{a}}.
\newblock Wavelet-based fourier information interaction with frequency diffusion adjustment for underwater image restoration.
\newblock In \emph{Proceedings of the IEEE/CVF Conference on Computer Vision and Pattern Recognition}, 8281--8291.

\bibitem[{Zhao et~al.(2024{\natexlab{b}})Zhao, Cai, Dong, and Zeng}]{10448182}
Zhao, C.; Cai, W.; Dong, C.; and Zeng, Z. 2024{\natexlab{b}}.
\newblock Toward Sufficient Spatial-Frequency Interaction for Gradient-Aware Underwater Image Enhancement.
\newblock In \emph{ICASSP 2024 - 2024 IEEE International Conference on Acoustics, Speech and Signal Processing (ICASSP)}, 3220--3224.

\bibitem[{Zheng and Wu(2024)}]{zheng2024ushapedvisionmambasingle}
Zheng, Z.; and Wu, C. 2024.
\newblock U-shaped Vision Mamba for Single Image Dehazing.
\newblock arXiv:2402.04139.

\bibitem[{Zhou et~al.(2024{\natexlab{a}})Zhou, Li, Ma, Yang, and Yang}]{zhou2024migcadvancedmultiinstancegeneration}
Zhou, D.; Li, Y.; Ma, F.; Yang, Z.; and Yang, Y. 2024{\natexlab{a}}.
\newblock MIGC++: Advanced Multi-Instance Generation Controller for Image Synthesis.
\newblock arXiv:2407.02329.

\bibitem[{Zhou et~al.(2024{\natexlab{b}})Zhou, Li, Ma, Zhang, and Yang}]{Zhou_2024_CVPR}
Zhou, D.; Li, Y.; Ma, F.; Zhang, X.; and Yang, Y. 2024{\natexlab{b}}.
\newblock MIGC: Multi-Instance Generation Controller for Text-to-Image Synthesis.
\newblock In \emph{Proceedings of the IEEE/CVF Conference on Computer Vision and Pattern Recognition (CVPR)}, 6818--6828.

\bibitem[{Zhou, Yang, and Yang(2023)}]{zhou2023pyramiddiffusionmodelslowlight}
Zhou, D.; Yang, Z.; and Yang, Y. 2023.
\newblock Pyramid Diffusion Models For Low-light Image Enhancement.
\newblock arXiv:2305.10028.

\bibitem[{Zhou et~al.(2024{\natexlab{c}})Zhou, Sun, Li, Jiang, Zhou, Lam, Zhang, and Fu}]{zhou2024hclr}
Zhou, J.; Sun, J.; Li, C.; Jiang, Q.; Zhou, M.; Lam, K.-M.; Zhang, W.; and Fu, X. 2024{\natexlab{c}}.
\newblock HCLR-net: Hybrid contrastive learning regularization with locally randomized perturbation for underwater image enhancement.
\newblock \emph{International Journal of Computer Vision}, 1--25.

\bibitem[{Zhu et~al.(2024)Zhu, Liao, Zhang, Wang, Liu, and Wang}]{zhu2024visionmambaefficientvisual}
Zhu, L.; Liao, B.; Zhang, Q.; Wang, X.; Liu, W.; and Wang, X. 2024.
\newblock Vision Mamba: Efficient Visual Representation Learning with Bidirectional State Space Model.
\newblock arXiv:2401.09417.

\end{thebibliography}
\end{document}